\documentclass[11pt,a4paper]{article}
\usepackage[hyperref]{emnlp-ijcnlp-2019}
\usepackage{times}
\usepackage{latexsym}
\usepackage{graphicx}
\usepackage{dirtytalk}
\usepackage{textgreek}
\usepackage{textcomp, xspace}

\usepackage{url}

\aclfinalcopy 

\title{Dependency Parsing for Spoken Dialog Systems}

\author{Sam Davidson\textsuperscript{1}, Dian Yu\textsuperscript{2}, Zhou Yu\textsuperscript{2} \\
 \textsuperscript{1}Department of Linguistics \\
 \textsuperscript{2}Department of Computer Science\\
 University of California, Davis \\
 Davis, CA 95616, USA \\
 {\tt \{ssdavidson,dianyu,joyu\}@ucdavis.edu} 
}

\date{}

\begin{document}
\maketitle
\begin{abstract}
Dependency parsing of conversational input can play an important role in language understanding for dialog systems by identifying the relationships between entities extracted from user utterances.
Additionally, effective dependency parsing can elucidate differences in language structure and usage for discourse analysis of human-human versus human-machine dialogs. 
However, models trained on datasets based on news articles and web data do not perform well on spoken human-machine dialog, and currently available annotation schemes do not adapt well to dialog data. Therefore, we propose the Spoken Conversation Universal Dependencies (SCUD) annotation scheme that extends the Universal Dependencies (UD) \cite{nivre2016universal} guidelines to spoken human-machine dialogs.  We also provide ConvBank, a conversation dataset between humans and an open-domain conversational dialog system with SCUD annotation. Finally, to demonstrate the utility of the dataset, we train a dependency parser on the ConvBank dataset. We demonstrate that by pre-training a dependency parser on a set of larger public datasets and fine-tuning on ConvBank data, we achieved the best result, 85.05\% unlabeled and 77.82\% labeled attachment accuracy.
\end{abstract}

\section{Introduction}
Syntactic parsing \cite{chengunrock} and semantic parsing \cite{amrl, semantic} have been used in dialog systems 
to disambiguate the relationships between noun phrases in understanding tasks. 
Compared to constituency parsing and semantic role labeling, dependency parsing provides more clear relationships between predicates and arguments \cite{vs_srl}. 

Constituency parsers provide information about noun phrases in a sentence, but provide only limited information about relationships within a noun phrase. For example, in the sentence \say{What do you think about Google's privacy policy being reviewed by journalists from CNN?,} a constituency parser would place \say{Google's privacy policy being reviewed by journalists from CNN} under a single phrasal node. Similarly, a semantic role labeling system would tend to label the same phrase as an argument of the verb, but it would not disambiguate the relationships within the phrase. Finally, NER only provides information about named entities which may or may not be the key semantic content of the sentence. Dependency parsers, by contrast, can provide information about relationships when a sentence contains multiple entities, even when those entities are within the same phrase. 

Identifying relationships between entities in a user utterance can help a dialog system formulate a more appropriate response. For instance, in the sentence about \say{Google's privacy policy} mentioned above, there are multiple entities for the system to consider.
The system must determine the most important entity in the utterance in order to model the topic and generate an appropriate response. A dependency graph facilitates this process by providing information about the syntactic relationships between entities.

\begin{figure}[h]
\includegraphics[width=7.7cm]{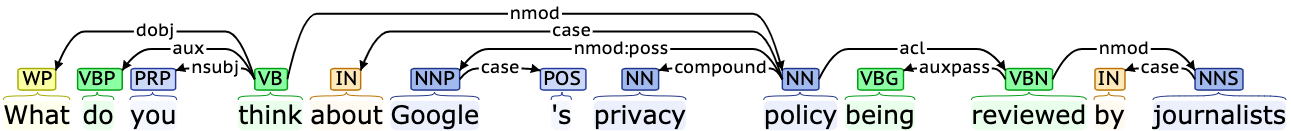}
\caption{Example dependency-parsed sentence. Stanford CoreNLP visualizer \cite{manning-EtAl:2014:P14-5}.}
\label{fig:journalists}
\end{figure}

A dependency parse of the sentence, shown in Figure \ref{fig:journalists}, demonstrates that the key element of the sentence, \say{policy}, is directly linked to the root \say{think} as a \textit{nmod}. The other entities in the sentence are modifiers of the central element, and are clearly labeled as such. These structures make it easier for a dialog system to identify core semantic content and formulate an appropriate response.

Dependency parsing is a well studied \cite{dozat2016deep, cvt} task. However, most dependency parsers are trained on well-written text datasets, such as news articles and web-scrape data. 
In a conversational setting, dialog systems deal with transcripts that are produced by automated speech recognition (ASR) systems. ASR systems are prone to produce errors, such as dropped and mis-transcribed words, and prematurely terminated utterances. Such errors, in our experience, make building an automated dependency parser harder.



Additional data is needed to train parsers for the dialog system domain. Due to limitations in the UD scheme for parsing speech data, we propose an extension of the UD standard, SCUD, designed specifically for dialog systems. SCUD considers unique issues in automated speech transcription and what is useful for dialog systems to generate appropriate responses. We then use SCUD to annotate a corpus of open-domain human-system conversations with automated speech transcripts. We finally leverage this data, along with previously annotated UD datasets, to build a dependency parser. Since the training data used is collected with an open-domain dialog system, the trained dependency parser can be applied to dialog systems with any task.

In addition to its use in dialog systems, dependency parsed dialog data could be used in discourse analysis to assist in understanding the syntactic patterns people use when interacting with a computer. While researchers have investigated such differences in task-oriented dialog systems \cite{doran2003comparing}, we believe that ConvBank and SCUD present an opportunity for expanded discourse analysis of open-domain dialog systems.

\section{Related Work}
Dependency parsing on text such as the Universal Dependencies English Web Treebank (UD-EWT) \cite{silveira14gold}, is a well-developed field in NLP; however, the parsing of automated speech transcripts remains an open problem \cite{bechet2014adapting}. As has been well-documented, speech, and the resulting ASR transcipts, is a very different domain from written text \cite{biber1999lexical, leech2000grammars, carter2017spoken}. 
\citet{Adams2017DependencyPA} focuses on adapting a parser trained on UD-EWT to the speech domain without annotating new training data. For example, she attempts to create additional speech domain training data by parsing raw speech transcripts with Google's SyntaxNet. She then uses the resulting parsed output as training data for MaltParser \cite{Nivre06maltparser:a}. However, the resulting models do not show consistent improvement over a baseline parser. Based on these results, and the apparent need for new training data for the speech domain, we chose to annotate a new human-system dialogs dataset to facilitate building dependency parsers for automated speech transcription and dialog systems.

Previous work has demonstrated that expanding the UD annotation scheme can result in successful parsers for other domains. For example, \citet{liu2018parsing} expand the UD scheme to train a parser for Twitter data. Thus we take a similar approach in expanding the UD scheme to encompass issues common to automated speech transcription.

Unlike written discourse, speech is full of disfluencies which make discovering the underlying syntactic structure challenging, as such disfluencies interrupt the syntactic structure of the utterance \cite{nasr2014automatically,caines2017parsing}. For example, according to \citet{meteer_dysfluency_1995}, 17\% of tokens in the Switchboard telephone conversations are various disfluencies. Frequent types of disfluencies include hesitations and false starts. When dealing with ASR-based systems, one must also contend with ASR errors. According to \citet{nasr2014automatically}, such disfluencies must either be integrated into the syntactic parse of the utterance or removed prior to parsing. We choose to take the former approach, integrating speech disfluencies and speech transcription errors into the dependency graph produced by our annotation scheme and the resultant parser.

\section{Annotation Scheme}
When we initially began to annotate speech transcript data using the UD 2.0 \cite{ud2} scheme, our annotators repeatedly complained of constructions which they were unable to annotate given the limitations of UD 2.0. We developed SCUD in response to these issues. SCUD closely follows UD 2.0 so that we may leverage larger UD-compliant datasets of text data, such as UD-EWT. We extend the UD annotation scheme to encompass specific features peculiar to the speech domain, namely the use of non-syntactic discourse markers, speech disfluencies, and ASR errors.

One major difference between speech and text is the fact that speech contains a fairly large number of discourse markers which do not serve a syntactic role in the utterance. This is similar to the difference between text and tweets pointed out by \citet{liu2018parsing}. We make use of the UD relation \textit{discourse} to indicate discourse markers which are not part of the syntactic structure of an utterance; for example, the word \say{like} in \say{I have like three dogs}, where \say{like} plays a pragmatic, rather than a syntactic, role. 

Another goal of our annotation scheme is to handle disfluencies common to both human and ASR-generated speech transcripts. We expand the scope of extant UD relation tags to tag common speech disfluencies. For example, the UD tag \textit{reparandum} is applicable to situations in which a user self-corrects his or her speech. We use the UD tag \textit{flat} to link repeated words and stutters to indicate that these headless multi-word expressions are a single syntactic unit. Handling non-syntactic tokens is particularly important because, while ASR may improve, the way people speak is less likely to change.

We are also able to expand the existing UD annotation scheme to handle some common errors in ASR output. In cases where an ASR system incorrectly splits words, we utilize the UD relation \textit{goeswith} to link the two words in a dependency graph. We also provide a means to insert nodes at locations where an ASR system has dropped words necessary for formation of a grammatical structure. We do this by expanding the enhanced UD 2.0 standard, which provides a method to insert nodes in graphs as place-holders for words omitted through ellipsis. An example of using such a place-holder node is shown in Table \ref{ellipsis}, where the subject of the sentence in question has been omitted due to an ASR error. We also create one new relation tag to deal with a common error in ASR system output: prematurely terminated user utterances. Many conversational agents are turn-based \cite{këpuska2018next} , and depend on pauses in user speech to detect the end of the user's dialog turn. As a result, hesitations by the user can result in the system prematurely terminating the user's turn. In reviewing our dialog system output data, we found that the ASR system prematurely terminates as many as 5\% of user turns. We created the dependency label \textit{preterm} to annotate such occurrences in our dataset.

\begin{table}[t!]
\begin{center}
\begin{tabular}{|l|l|l|l|l|}
\hline \multicolumn{5}{|l|}{\bf{Transcript: got two dogs}} \\ \hline
\bf Index & \bf Word & \bf POS & \bf Dep. & \bf Relation \\ \hline
1 & E1.1 & PRON & 2 & nsubj \\
2 & got & VERB & 0 & root \\
3 & two & NUM & 4 & nummod \\
4 & dogs & NOUN & 2 & obj \\
\hline
\end{tabular}
\end{center}
\caption{\label{ellipsis} Example of omitted word node insertion }
\end{table}

We must also consider the issue of incorrectly transcribed words. ASR systems are prone to mistranscribing words in the input signal; such errors can significantly change the meaning of the underlying utterance or result in a transcription which is ungrammatical or semantically nonsensical \cite{errattahi2018improving}. While error correction methods for dialog systems have been developed \cite{Choi2016}, preprocessing our data with such a system is beyond the scope of this study. We instructed annotators to annotate all words as they were transcribed by the system. The alternative would involve annotators trying to infer user intent from context, which makes predictable annotation far less likely. 

In total, SCUD has 37 primary relation tags. The relation tagset closely follows the UD 2.0 annotation standards with the exception of domain-specific extensions as noted above.

\section{Dataset and Annotation Process}

We collected a corpus of humans conversing with an open-domain dialog chatbot based on the Alexa platform \cite{chengunrock}. 
Users were asked to come into a lab and have an undirected conversation with our system. The automated speech transcripts of the users' utterances were collected and selected at random for inclusion in this corpus. We recruited two annotators who are experienced with dependency grammars and annotation projects to annotate 1,500 sentences. As is common with other UD annotation projects, we asked each annotator to correct the work of the other. Once this round of corrections was complete, we measured agreement as the percentage of tokens which remained unchanged between the original annotations and the corrected version. The two annotators agreed on 86.4\% of unlabeled dependencies and 80.5\% of labeled dependencies. These numbers are comparable to those reported on Tweebank \cite{liu2018parsing}, another challenging annotation problem. Once annotator agreement was assessed, a third annotator reviewed the annotations to improve compliance with the annotation scheme. We found that ASR errors, such as ungrammatical sentences resulting from mistranscription and word omission are a major contributing factor to annotator disagreement in our dataset.


The result of our annotation project is ConvBank, a corpus of 1,500 ASR-generated text utterances from a human-machine spoken dialog system with SCUD annotation. The corpus is compliant with UD 2.0 standards with the exception of extensions made to the UD standard as noted above. All data, code, and pre-trained models related to this project are publicly available via the project webpage\footnote{https://gitlab.com/ucdavisnlp/dialog-parsing}.


ConvBank uses 32 of the 37 possible primary relations in SCUD, described above. A comparison of the ten most common tags in ConvBank and UD-EWT is presented in Table \ref{rel-tags}. \textit{punct} is the most common relation in UD-EWT but is absent from ConvBank; this is because our automated speech transcripts do not contain punctuation. Other differences in relation distribution, such as the higher relative frequencies of \textit{root} and \textit{nsubj} can be explained by the shorter sentences in ConvBank compared to UD-EWT. Since each sentence must have a root, and most have a subject, the fewer other words in each sentence, the higher the frequencies of these two tags will tend to be.

\begin{table}[t!]
\begin{center}
\begin{tabular}{|l|r||l|r|}
\hline \multicolumn{2}{|c||}{\bf{UD-EWT}} & \multicolumn{2}{|c|}{\bf{ConvBank}} \\ \hline
\bf Tag & \bf Freq. & \bf Tag & \bf Freq. \\ \hline
punct & 11.5\% & root & 15.1\% \\
case & 8.5\% & nsubj & 13.8\% \\
nsubj & 8.0\% & obj & 9.2\% \\
det & 7.7\% & advmod & 8.2\% \\
root & 6.1\% & case & 7.2\% \\
advmod & 5.4\% & det & 6.8\% \\
obj & 5.0\% & obl & 6.0\% \\
obl & 4.5\% & aux & 5.3\% \\
amod & 4.4\% & cop & 4.1\% \\
compound & 4.0\% & amod & 3.2\% \\
\hline
\end{tabular}
\end{center}
\caption{\label{rel-tags} Comparison of common relation frequencies. }
\end{table}

\section{Parsing Experiments}

To demonstrate the utility of the ConvBank dataset, we use the data to train a dependency parser for ASR-based spoken dialog systems. 

\subsection{Training Data}
We use three datasets in training our dependency parser. We use the publicly available UD-EWT \cite{silveira14gold} and Tweebank v.2 \cite{liu2018parsing} to train baseline models. We also combine UD-EWT and Tweebank to create a larger UD dataset for pre-training. Finally, we use our own ConvBank dataset to fine-tune our best-performing baseline model. We hold out 500 annotated ConvBank sentences for testing.

We considered using the constituency parsed Switchboard corpus \cite{Godfrey:1992:STS:1895550.1895693}, which is a human-human telephone corpus, due to its similarity to our domain. However, no fully UD-compliant version of the dependency parsed corpus is publicly available. Although conversion from constituency to dependency graphs is possible, we found the resulting dependency graphs unusable due to the error-prone conversion process. \citet{peng2018all} point out that conversion systems particularly struggle with attaching nominal modifiers and oblique arguments. 

\subsection{Methods}
We build all models using the biaffine parser as described in \citet{dozat2016deep} and \citet{qi2019universal}. We use the open-source implementations of Dozat's biaffine parser code\footnote{https://github.com/tdozat/Parser-v3} for training our models. We use Glove \cite{glove} for pre-trained word embeddings.
We train three baseline models to compare how models trained on text data compare with our model, which leverages both annotated text and automated speech transcripts. First, we train a parsing model on UD-EWT \cite{silveira14gold} alone. We also train a separate model on Tweebank v.2 \cite{liu2018parsing} alone to determine how a parser trained on tweets, which are arguably more similar to conversational speech, will perform. We also train a model on combined UD-EWT and Tweebank data. Our final baseline is a model trained on a 1,000 utterance subset of ConvBank. This model is used to determine how a parser trained on our data performs without pre-training on a larger text dataset.

Our fine-tuning approach is based on the idea that, by providing the system with additional domain-specific examples from ConvBank, we are able to build upon the system's baseline knowledge to create a final system which performs substantially better than baseline. \citet{stymne2018parser} demonstrate that this type of fine-tuning is an effective way to leverage multiple treebanks to train a single dependency parser. Our fine-tuning process consists of initializing our network weights using the weights from our baseline model trained on the UD English Web Treebank. As demonstrated by our results, the baseline model trained on UD-EWT alone performs reasonably well in assigning many UD labels and dependencies; however the system does not handle well issues specific to the dialog domain. To improve performance on the dialog domain, we resume training the pre-trained model with our annotated ConvBank data, using identical training parameters, until perplexity stabilizes.


\subsection{Results and Discussion}
We present the parsing accuracy on the ConvBank test set in Table \ref{results}. Training only using 1,000 annotated in-domain ConvBank data reaches 75.65\% and 58.29\% accuracy on unlabeled and labeled edges respectively. This suggests that 1,000 utterances are not enough to achieve optimal results. However, human annotations are expensive to obtain, limiting the size of our dataset. The model trained on EWT or Tweebank or EWT and Tweebank combined achieved similar results that are better than the model trained on ConvBank alone. As these three models had similar results, we fine-tune on the slightly better-performing EWT model. Fine-tune achieves the best best result on both unlabeled and labeled edges (85.05\% and 77.82\% respectively). Fine-tuning on EWT with ConvBank data improves parsing accuracy on the test data by over 7 points on unlabeled dependency edges and by nearly 12 points on labeled edges. These results suggest leveraging both a large set of out-of-domain data and in-domain data achieves the best performance. 

There are some clear differences between our fine-tuned model and the baseline model trained on text data. In speech, discourse markers such as \say{but}, \say{so}, and \say{and} frequently begin utterances. These items can be difficult to parse correctly, as they more frequently serve grammatical functions. Our baseline model generally tags such items as coordinating conjunctions or adverbs, while in the fine-tuned model such items are more often appropriately tagged as discourse markers. Another clear example is the correct use of the \textit{reparandum} tag. For example in the utterance \say{you know you my name}, the baseline model tags the second \say{you} as an indirect object, while the fine-tuned model correctly tags the item as \textit{reparandum}. The baseline model tags only three items in our test data as \textit{reparandum}, while the fine-tuned model identifies thirty-one such instances. These results indicate that our fine-tuned model is more closely aligned to the speech domain than our baseline model. However, our best model still struggles when faced with fixed expressions common in conversational speech, such as \say{Top Gun} and \say{Kitchen Table Wisdom}. The model also performs poorly on ungrammatical utterances, as expected.

\begin{table}[t!]
\begin{center}
\begin{tabular}{|l|r|r|}
\hline \bf Dataset & \bf UAS & \bf LAS \\ \hline
\bf UD-EWT & 77.41 & 65.85 \\
\bf Tweebank & 77.23 & 66.44 \\
\bf Tweebank\texttt{+}UD-EWT & 77.38 & 65.75 \\
\bf ConvBank & 75.65 & 58.29 \\
\bf Fine-tune 
& 85.05 & 77.82 \\
\hline
\end{tabular}
\end{center}
\caption{\label{results} Parsing Accuracy Results}
\end{table}

\section{Conclusion}
Dependency parsing can effectively identify the relationships between entities in a user utterance to improve speech language understanding. However, dependency parsing for speech data does not perform well when trained on out-of-domain text data. To solve this problem, we present ConvBank, a new dataset of annotated automated speech transcripts, 
annotated with the Spoken Conversation Universal Dependencies (SCUD) annotation scheme to build UD-compatible parses. 
We demonstrate the utility of ConvBank by fine-tuning a dependency parser with a large dataset, reaching 85.05\% on labeled dependency edges and 77.82\% on unlabeled dependencies.

In future work, we hope to expand the size of ConvBank to improve its utility to researchers in training parsing models and studying conversational systems. We also hope to improve the performance of our parsing system by jointly training dependency parsing with semantic role labeling.

\section*{Acknowledgements}
We would like to thank our skilled annotators, Hekang Jia and Julian Rambob.

\bibliography{emnlp-ijcnlp-2019}
\bibliographystyle{acl_natbib}
\end{document}